\pdfoutput=1

\documentclass[11pt]{article}

\usepackage{naacl2021}

\usepackage{times}
\usepackage{latexsym}

\usepackage[T1]{fontenc}

\usepackage[utf8]{inputenc}

\usepackage{microtype}

\usepackage{caption}
\usepackage{subcaption}
\usepackage{graphicx}
\usepackage{multirow}
\usepackage{multicol}
\usepackage{amsmath}
\usepackage{amssymb}
\usepackage[para]{threeparttable}
\usepackage{booktabs}
\usepackage{tabularx}
\usepackage{makecell}
\usepackage{algorithm}
\usepackage{algorithmic}
\usepackage{pifont}
\usepackage{mathabx}
\usepackage{xcolor}
\usepackage{mathtools}

\DeclarePairedDelimiter\floor{\lfloor}{\rfloor}

\title{Designing a Minimal Retrieve-and-Read System \\for Open-Domain Question Answering}

\author{
    Sohee Yang\textnormal{\textsuperscript{1}}\thanks{\;\, Most of the work was done while the author was working at NAVER Corp.} \and Minjoon Seo\textnormal{\textsuperscript{1,2}} \\
    \textsuperscript{1}KAIST AI \quad \textsuperscript{2}NAVER Corp. \\
    \texttt{\{sohee.yang,minjoon\}@kaist.ac.kr}
}

\begin{document}
\maketitle
\begin{abstract}

In open-domain question answering (QA), retrieve-and-read mechanism has the inherent benefit of interpretability and the easiness of adding, removing, or editing knowledge compared to the parametric approaches of closed-book QA models.
However, it is also known to suffer from its large storage footprint due to its document corpus and index.
Here, we discuss several orthogonal strategies to drastically reduce the footprint of a retrieve-and-read open-domain QA system by up to 160x.
Our results indicate that retrieve-and-read can be a viable option even in a highly constrained serving environment such as edge devices, as we show that it can achieve better accuracy than a purely parametric model with comparable docker-level system size.\footnote{Our code and model weights are available in \\\href{https://github.com/clovaai/minimal-rnr-qa}{https://github.com/clovaai/minimal-rnr-qa}.}

\end{abstract}

\section{Introduction}

Open-domain question answering (QA) is the task of finding answers to generic factoid questions. In recent literature, the task is largely approached in two ways, namely \emph{retrieve \& read} and \emph{parametric}. The former solves the problem by first retrieving documents relevant to the question from a large knowledge source and then reading the retrieved documents to find out the answer~\citep{orqa, realm, dpr, rag, fid}. The latter, also known as closed-book QA, generates the answer in a purely parametric end-to-end manner~\citep{gpt3, t5}.

\begin{figure}[t] 
\centering
\includegraphics[width=0.48\textwidth]{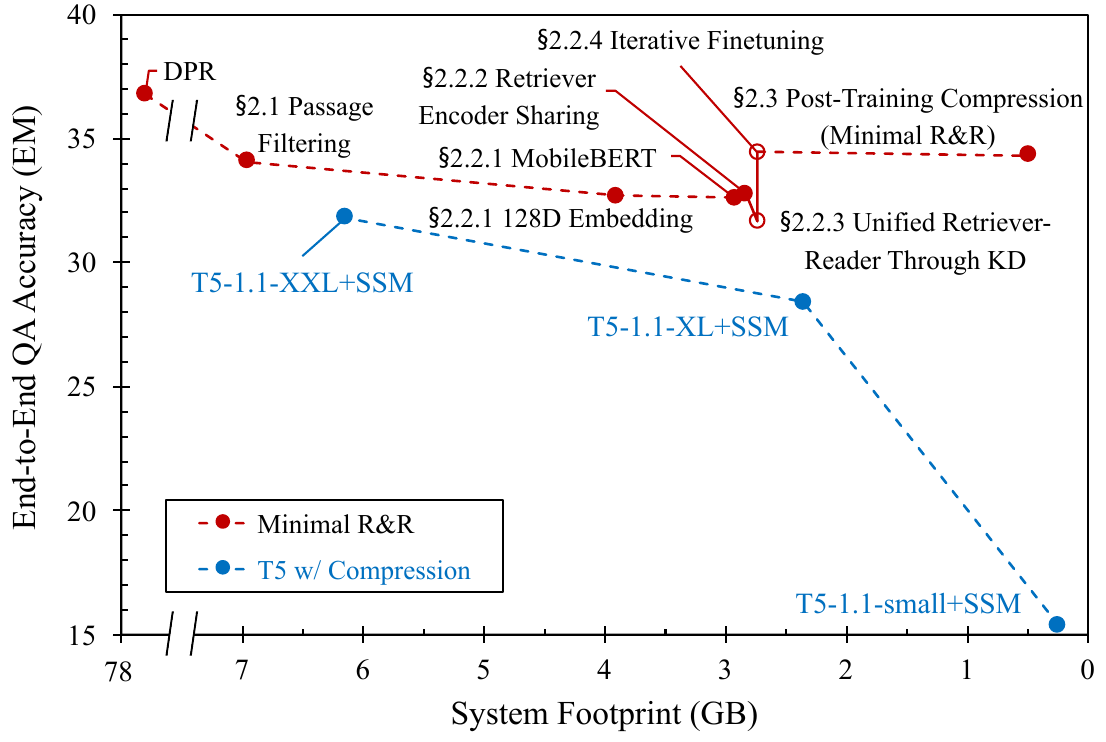}
\caption{\small System footprint vs. Exact Match (EM) accuracy on EfficientQA dev set.
System footprint is measured by the command \texttt{du -h /} inside the standalone docker container as stated in the EfficientQA competition guideline.
The red plot from left to right shows a path of reducing the size of an open-domain QA system with DPR from 77.5GB to 484.68MB by successively applying each of the strategies in Section~\ref{sec:method}.
The storage footprints of the baseline systems with T5 are calculated assuming the use of the lightweight docker image and post-training compression methods applied to our system.
}
\label{fig:main}
\end{figure}

While a parametric model enjoys the benefit in terms of system size that they do not require additional knowledge source like a retrieve \& read system does, its fundamental limitations are that their predictions are not so interpretable and they are not suitable for dynamic knowledge source as it is difficult to add, remove, or edit knowledge in the parametric model. These limitations are well-addressed by the retrieve \& read mechanism, which makes it often more suitable for real-world products. However, it is known to suffer from its large storage footprint due to its document corpus and index, especially compared to the parametric model that only needs to store the parameters~\citep{memory, r2d2, paq}.

Building an interpretable and flexible open-domain QA system and reducing its system size are both important in real-world scenarios; the system must be able to quickly adapt to the changes of the world and be deployed in a highly constrained serving environment such as edge devices. Hence, to get the best of both worlds, it is worthwhile to explore the trade-off between the storage budget and the accuracy of a retrieve \& read system.

Well-known approaches for reducing the size of a neural network include pruning~\citep{pruning}, quantization~\citep{q8bert}, and knowledge distillation~\citep{distillation}. In this paper, we utilize some of these generic approaches and combine them with problem-specific techniques to size down a conventional retrieve \& read system. We first train a passage filter and use it to reduce the corpus size (Section~\ref{sec:filtering}). We further apply parameter sharing strategies and knowledge distillation to make a single-encoder lightweight model that can perform both retrieval and reading (Section~\ref{sec:sharing}). In addition, we adopt multiple engineering tricks to make the whole system even smaller (Section~\ref{sec:compression}).

We verify the effectiveness of our methods on the dev set and test set of EfficientQA\footnote{A recently-hosted open-domain QA challenge at NeurIPS 2020 in which 18 teams participated. \\ \href{https://efficientqa.github.io}{https://efficientqa.github.io}}~\citep{efficientqa}. By applying our strategies to a recent extractive retrieve \& read system, DPR~\citep{dpr}, we reduce its size by 160x with little loss of accuracy, which is still higher than the performance of a purely parametric T5~\citep{t5} baseline with a comparable docker-level storage footprint. In Appendix~\ref{appendix:nq_tqa}, we also report the performance on two more open-domain QA datasets, Natural Questions~\citep{nq} and TriviaQA~\citep{tqa}, to test the generalizability of our methods and suggest a future research direction.

\section{Method}\label{sec:method}

In this section, we discuss three techniques for reducing the storage footprint of a generic retrieve \& read system, namely passage filtering (Section~\ref{sec:filtering}), unifying retriever and reader into a single model through parameter sharing (Section~\ref{sec:sharing}), and post-training compression (Section~\ref{sec:compression}).
We assume that the initial system takes the conventional composition of a trainable (neural) retriever with a question encoder and a passage encoder that create dense vectors used for search, a neural extractive reader (possibly with passage ranking), and a text corpus and the corresponding index that serve as the knowledge source.
Figure~\ref{fig:main} shows how we start from one such retrieve \& read system and apply each of the methods, in the order they are introduced in this section, to successively reduce its system footprint without sacrificing much accuracy.

\subsection{Passage Filtering}\label{sec:filtering}
Index and corpus files can take up a significant portion of the storage footprint of a retrieve \& read system if a large text corpus is utilized as the knowledge source.
Therefore, to drastically reduce the system size, we train a binary classifier and use it to exclude passages that are relatively unlikely to be useful for question answering.

Let the set of indices of all passages in the corpus be $J_\text{total}$. To create the training data, we split $J_\text{total}$ into two disjoint sets, $J^+_\text{train}$ and $J^-_\text{train}$, such that the former contains the indices of the passages we would like to include in the minimal retrieve \& read system.\footnote{All the details, including how we split the data, are in Appendix~\ref{appendix:filter}.}
Denoting $E(\cdot)$ as a trainable dense encoder which maps a passage to a $d$-dimensional embedding, such that $v_j = E(p_j) \in \mathbb{R}^{d}$, the score $s_j = v_j^\top w$, where $w \in \mathbb{R}^{d}$ as a learnable vector, represents how likely a passage $p_j$ would hold the answer to an input question.
The classifier is trained with binary cross entropy on the minibatches of half-positive and half-negative passages, $J^{+'}_\text{train}$ and $J^{-'}_\text{train}$ drawn from $J^+_\text{train}$ and $J^-_\text{train}$, respectively.

During training, we sample several checkpoints and evaluate them using the hit ratio on a validation set: $\text{hit}_\text{val} = |J^{[1:|J_\text{val}^+|]}_\text{val} \cap J^+_\text{val}| / |J^+_\text{val}|$, where $J^+_\text{val}$ is a set of indices of the ground truth passages that hold the answer for the questions in the validation set and $J^{[1:|J_\text{val}^+|]}_\text{val}$ is the set of indices $j$ of the passages whose inferred score $s_j$ is in the top-$|J_\text{val}^+|$ scores sorted in descending order, among all $s_j$ such that $j \in J^+_\text{val} \cup J^-_\text{val}$. $J^-_\text{val}$ is a disjoint set randomly sampled from $J^{-}_\text{train}$.

We select the checkpoint with the highest $\text{hit}_\text{val}$ and calculate $s_j$ for all $p_j$, where $j \in J_\text{total}$, using the selected checkpoint. Then, we retrieve $J_\text{subset} = J^{[1:n]}_\text{total}$, the set of indices of the $n$ top-scoring passages, to indicate the passages to include in our minimal retrieve \& read system.

\subsection{Retriever-Reader with Single Encoder}\label{sec:sharing}

In this subsection, we introduce how to obtain a unified retriever-reader with a single encoder (which results in a smaller system footprint) that can perform both retrieval and reading without much drop in accuracy. The unified retriever-reader is trained by successively applying (1) retriever encoder sharing, (2) distilling a reader into the retriever-reader network, and (3) iterative finetuning.

\paragraph{2.2.1 Lightweight Encoder and Embedding Dimension Reduction}\label{sec:lightweight}
To make the system small, we utilize a lightweight pretrained encoder. Specifically, MobileBERT~\citep{mobilebert} (4.3x smaller than BERT-base~\citep{bert}) is employed as the encoder of our retriever-reader model.

We use the dense embedding vectors of the passages in the knowledge source as the index. Therefore, reducing the embedding dimension results in a linear decrease in the index size. We use only the first 128 dimensions (out of 512) to encode the questions and passages.

\paragraph{2.2.2 Retriever Encoder Sharing}
Let $E_\psi(\cdot)$ and $E_\phi(\cdot)$ be the the question encoder and passage encoder of a retriever, where each of the encoders produces a vector for question $q$ and passage $p$.

We share the parameters of the encoders, so that $\psi = \phi = \theta$, and differentiate the question inputs from passages inputs using an additional input signal: different token type ids of 0 for questions and 1 for passages. The retrieval score for a pair of question $q$ and passage $p$ is calculated as $\text{sim}_\theta(q, p) = E_\theta(q, 0) \cdot E_\theta(p, 1)$.

We minimize the negative log-likelihood of selecting the passage which holds the answer, namely the positive passage, while training on mini-batches that consist of questions that are each paired with one positive passage and several negative passages. This procedure creates a retriever with a single encoder of parameters $\theta$ that can encode both questions and passages.\footnote{In a setting where the index is frozen (addition or editing of index items does not occur), the system does not need a passage encoder. However, we assume a self-contained system with the full ability to update the index, so the passage encoder is considered in the system composition.}

\paragraph{2.2.3 Unified Retriever-Reader Through Knowledge Distillation}

The previous subsection describes how to make a retriever that holds only one encoder. Here, we further train the parameters of the retriever $\theta$ so that it can also acquire the ability of a reader; we make a unified retriever-reader model that shares all the encoder parameters and eliminate the need for a separate reader. Specifically, using a fully trained reader of parameters $\xi$ as the teacher, we adopt knowledge distillation to transfer its reading ability to the unified retriever-reader network. The training starts after initializing the parameters of the retriever-reader as $\theta$, which is obtained from the retriever encoder sharing procedure described in the previous subsection.

Let $J_\text{read} \subset J_\text{subset}$ be the set of indices of the passages whose retrieval score $\text{sim}_\omega(q, p_j)$, calculated for question $q$ using a retriever with parameters $\omega$\footnote{The retriever with parameters $\omega$ is the retriever used with the teacher reader of parameters $\xi$.}, is among the top-$k'$ scores for all $j \in J_\text{subset}$.
$J_\text{read}$ serves as the candidate pool of the indices of the training set passages.

During training, for question $q$, a set of passages $P_q = \{p_i | 1 \leq i \leq m \}$ where $m \geq 2$ is sampled from $\{p_j | j \in J_\text{read}\}$ to construct a part of the training batch, such that only $p_1$ contains the answer to question $q$ among $p_i \in P_q$.

Then, we train the unified retriever-reader network with parameters $\theta$ using a multitask loss $\mathcal{L}_\text{read} + \mathcal{L}_\text{ret}$, such that the former is used to train the reader part of the network, and the latter is used to keep training the retriever part. The resulting retriever-reader model has the ability to perform both retrieval and reading.

$\mathcal{L}_\text{read}$ is designed to distill the knowledge of a reader teacher into the reader part of the retriever-reader student; the KL divergence between the sharpened and softmaxed answer span scores of the teacher and the student, $D_{KL}(\mathcal{P}^\text{span}_{\xi, q} || \mathcal{P}^\text{span}_{\theta, q})$. If the teacher reader additionally contains a passage ranker, distillation is also jointly done on the passage ranking scores ($m$-dim vector outputs).

Retrieval loss $\mathcal{L}_\text{ret}$ is jointly optimized in a multitask-learning manner to prevent the retriever part of the unified network from forgetting the retrieval ability while training the reader part. The loss can either be the negative log-likelihood described in the previous subsection or another knowledge distillation objective function with a fully trained retriever teacher. If the reader teacher used for $\mathcal{L}_\text{read}$ has a passage ranker, the passage ranking score of the teacher can serve as the distillation target~\citep{rdr}.

\paragraph{2.2.4 Iterative Finetuning of Unified Retriever-Reader}

We have observed that finetuning the unified retriever-reader for a few more epochs leads to better retrieval and reading performance. While the most simple method is to jointly train the model on the standard reader loss and retriever loss\footnote{The marginal negative log-likelihood of all the correct answer spans in the positive passage and the negative log-likelihood of positive passage $p_1$ being selected, respectively.}, we additionally try iterative finetuning of each of the retriever and reader part as described in Algorithm~\ref{alg:finetuning}. The motivation here is to apply a loose reconstruction constraint $\mathcal{L}_\text{recon}$ to keep the retrieval score as it is before and after the model is optimized for reading, with an assumption that this would be helpful to alleviate the train-inference discrepancy in the input distribution of the reader, created because the unified retriever-reader is not trained using a pipelined manner (training the reader on top of the retrieval result of a fixed retriever).

\begin{algorithm}[ht]
\small
   \caption{\small A single iterative finetuning step on the unified retriever-reader with parameters $\theta$ at time $t$}
   \label{alg:finetuning}
   
   \hspace*{\algorithmicindent} \textbf{Input}
    $\theta^{(t)}$ (parameters of the model at time $t$), knowledge distillation temperature $\tau$, and training batch of question $q$ and passages $P_q = \{p_i | 1 \leq i \leq m \}$ drawn from $J_\text{read}$ such that $m \geq 2$, $Y(q, p_1) = 1$, and $Y(q, p_i) = 0, \forall 2 \leq i \leq m$. (batch size of 1 is assumed here for a simple presentation) \\
    \hspace*{\algorithmicindent} \textbf{Output}
    Updated parameters $\theta^{(t+1)}$ \\
\begin{algorithmic}[1]
\STATE $\ell^{(t)} \gets [ E_\theta^{(t)} (p_1, 1), \cdots, E_\theta^{(t)} (p_m, 1) ]^\top E_\theta^{(t)} (q, 0)$
\STATE $\hat{\theta}^{(t)} \gets \text{GradientUpdate} ( \mathcal{L}_\text{read}( q, p_1, \cdots, p_m ); \theta^{(t)} )$
\STATE $\hat{\ell}^{(t)} \gets [ E_{\hat{\theta}}^{(t)} (p_1, 1), \cdots, E_{\hat{\theta}}^{(t)} (p_m, 1) ]^\top E_{\hat{\theta}}^{(t)} (q, 0)$
\STATE $\mathcal{L}_\text{recon} \gets D_\text{KL} \left( \text{softmax} (\ell^{(t)} / \tau) || \text{softmax} (\hat{\ell}^{(t)} / \tau) \right) $
\STATE $\mathcal{L}_\text{nll} \gets \text{CrossEntropy} ( \text{softmax} (\hat{\ell}^{(t)}), Y )$
\STATE $\theta^{(t+1)} \gets \text{GradientUpdate} ( \mathcal{L}_\text{recon} + \mathcal{L}_\text{nll} ; \hat{\theta}^{(t)} )$
\end{algorithmic}
\end{algorithm}
\vspace*{-.4cm}

\subsection{Post-Training Compression Techniques}\label{sec:compression}

In addition to the training methods to decrease the corpus, index, and model size, several post-training engineering tricks are applied to compress the system footprint further: (1) INT8 quantization of index items, (2) saving model weights as FP16, (3) resource compression, and (4) utilizing token IDs as the corpus instead of raw texts.

\paragraph{INT8 Quantization of Index Items}
The dense embeddings that serve as the items in the search index are of type FP32 in the default state. INT8 quantization can be applied to reduce the index size by four times with a little bit of drop in the accuracy. We make use of the quantization algorithm implemented in FAISS~\citep{faiss} IndexScalarQuantizer\footnote{\href{https://github.com/facebookresearch/faiss/blob/v1.5.2/\\IndexScalarQuantizer.cpp}{https://github.com/facebookresearch/faiss/blob/v1.5.2/\\IndexScalarQuantizer.cpp}}. During inference, the embeddings are de-quantized, and the search is performed on the restored FP32 vectors.

\paragraph{Saving Model Weights as FP16}
Half precision can be used to size down the model weights of originally FP32 tensors with almost no drop in accuracy. In PyTorch, this can be done by calling \texttt{.half()} on each FP32 tensor in the model checkpoint.

In TensorFlow, model graphs saved as the data type of FP16 may result in unacceptably slow inference according to the used hardware. We have found out that keeping the tensor types of the graph as FP32 but making the actual assigned values as FP16 enables a higher compression ratio when the model weights are compressed as described below.

\paragraph{Resource Compression}
Data compressors with a high compression ratio are effective at reducing the initial system footprint. Our observation is that bzip2 is better for binary files such as model weights or index of embedding vectors, whereas lzma is better for human-readable text files. System resources can also be compressed if necessary. We use \texttt{-9} option for both compressors.

\paragraph{Utilizing Token IDs as the Corpus}
A corpus file must be included in the system to get the actual text of the item retrieved by search (an embedding vector in our case). We have found out that using the file of the encoded token ids of the tokenized texts as the corpus, instead of the raw texts, is beneficial not only because it reduces the inference latency by preprocessing the texts, but also the compressed output size is often slightly smaller.

\section{Experiments}\label{sec:experiments}

\paragraph{Experimental Setup}
We apply our storage reduction methods to a recent extractive retrieve \& read system, DPR~\citep{dpr}, which consists of three different BERT-base encoders: question encoder of the retriever, passage encoder of the retriever, and encoder of the reader with a ranker.
All experiments are done on Naver Smart Machine Learning (NSML) Platform~\citep{nsml1, nsml2}.
The training and evaluation details are in Appendix~\ref{appendix:filter}, \ref{appendix:sharing}, and \ref{appendix:eval}.

\paragraph{Experimental Results}
Figure~\ref{fig:main} shows how each of the discussed strategies changes DPR's system size and Exact Match (EM) score on the EfficientQA dev set (see Table~\ref{table:ablations} and Table~\ref{table:overall} in Appendix for details). Our starting point is a standalone open-domain QA system with DPR whose estimated size is 77.5 GB: 1.4 (system) + 0.8 (retriever) + 0.4 (reader) + 61 (index) + 13 (text) GB. The red plot shows from left to right one path to successively apply each strategy to reduce the system footprint to 484.69MB, which is 160 times smaller.
Although the methods are described as sequential for easier presentation, the methods with filled markers and dotted lines are orthogonal to each other and thus can be applied in any other order. The methods with unfilled markers and solid lines are built on top of the previous method for each.

Sizing down the corpus from 21,015,325 to 1,224,000 (5.8\%) passages (\S 2.1) decreases the system footprint by a large margin of about 70.5GB with only 2.72\% of drop in EM.
Using a smaller passage embedding dimension of 128D (\S 2.2.1), changing the encoder to MobileBERT (\S 2.2.1), and sharing the encoders of the retriever (\S 2.2.2) save further 4.1GB of storage with little drop in accuracy of 1.28\%. The process of unifying the retriever and reader into a single model (\S 2.2.3) drops EM by 1.11, but the accuracy increases by 2.77\% (to 34.44\%) with iterative finetuning (\S 2.2.4).
In ablation studies on the three-step training procedure, omitting the knowledge distillation step drops EM by 1.5\%, and omitting $\mathcal{L}_\text{recon}$ drops EM by 0.38\%.

Applying post-training compression techniques further reduces the system footprint by a large margin while sacrificing little accuracy. EM changes to 34.39\% with INT8 quantization, and the rest of the tricks do not affect the accuracy. Converting the PyTorch checkpoint to a binary for TensorFlow Serving to reduce system library dependency and applying bzip2 compression on some of the system resources creates the final system of 484.69MB with an accuracy of 34.33\%.
Figure~\ref{fig:main} shows that this accuracy is higher than the performance of the parametric T5~\citep{t5} baseline with a comparable docker-level system footprint.\footnote{The accuracy of the T5 baselines are calculated using the SSM models finetuned on Natural Questions: \href{https://github.com/google-research/google-research/tree/master/t5_closed_book_qa\#released-model-checkpoints}{https://github.com/google-research/google-research/tree/ master/t5\_closed\_book\_qa\#released-model-checkpoints}.}

In Table~\ref{table:baseline}, we show the test set accuracy of our final system and other baselines. In summary, the performance of our system is higher than all of the parametric baselines, and the accuracy drop from DPR is only 2.45\% on the EfficientQA dev set and about 4\% on the test set while reducing the system footprint to about 0.6\% of the original size.

Our final system achieves the first place in the human (manual) evaluation and the second place in the automatic evaluation on ``Systems Under 500MB Track'' of the EfficientQA competition. While the accuracy of our system is 32.06\% on the EfficientQA test set in the automatic evaluation, which is 1.38\% behind the top-performing system~\citep{paq}, its accuracy is 42.23\% in the human evaluation which is 2.83\% higher than the other system. Interestingly, when possibly correct answers are also counted as correct, the accuracy rises to 54.95\% (7.58\% higher than the other system). Please refer to Table 2 of \citet{efficientqa} for more details.

In addition to the EfficientQA dataset, we also perform experiments on open-domain Natural Questions (NQ)~\citep{nq} and TriviaQA~\citep{tqa} to test the generalizability of the proposed methods. The results and detailed analysis are presented in Appendix~\ref{appendix:nq_tqa}.

\begin{table}[t]
    \setlength{\tabcolsep}{3.5pt}
    \centering
    \caption[Caption]{\small System size and Exact Match (EM) score of several standalone open-domain QA systems, reported on the test set of EfficientQA. The value of the baseline systems except for DPR are reported in the EfficientQA leaderboard\footnotemark. The system size of DPR is estimated as described in Section~\ref{sec:experiments}.}
    \label{table:baseline}
    \footnotesize
    \begin{threeparttable}
    \begin{tabular*}{0.48\textwidth}{l@{\extracolsep{\fill}}crc}
        \toprule
\textbf{Model}                     & \textbf{EM} & \textbf{System Size} & \textbf{Mechanism} \\
\midrule
T5-1.1-small+SSM & 18 & 486.61 MB & parametric       \\
T5-1.1-XL+SSM    & 28 & 5.65 GB   & parametric       \\
REALM            & 35 & 27.19 GB  & retrieve \& read \\
\midrule
DPR              & 36 & 77.5 GB   & retrieve \& read \\
+ Our Methods     & 32 & \textbf{484.69 MB} & retrieve \& read \\
        \bottomrule
        \vspace*{-.4cm}
    \end{tabular*}
    \end{threeparttable}
\end{table}
\footnotetext{\href{https://ai.google.com/research/NaturalQuestions/efficientqa}{https://ai.google.com/research/NaturalQuestions/ \\efficientqa}}
\vspace{-0.2em}

\section{Related Works}

There has recently been a line of work that targets to create storage-efficient open-domain QA systems, especially following the EfficientQA competition. Here, we introduce several approaches concurrent to ours that interested readers may refer to. \citet{memory} and \citet{r2d2} explore the trade-off between storage budget and accuracy, and their retrieve \& read systems take up only about 6GB with state-of-the-art performance. \citet{paq} propose a QA-pair retrieval system for open-domain QA, which enjoys the benefits of high flexibility and low latency.
Their retriever answers 1100 questions per second with 41.2\% accuracy on NQ, which rises to 47.7\% when equipped with a reranker. The variants optimized for small system footprint are the winning systems of two storage-constrained tracks at EfficientQA. \citet{efficientqa} review the EfficientQA competition with detailed analysis and summarize all of the top-performing systems.

\section{Conclusion}
We discuss several orthogonal approaches to reduce the system footprint of a retrieve-and-read-based open-domain QA system. The methods together reduce the size of a reference system (DPR) by 160 times with an accuracy drop of 2.45\% and 4\% on EfficientQA dev and test, respectively.
We hope that the presented strategies and results can be helpful for designing future retrieve-and-read systems under a storage-constrained serving environment.

\section*{Acknowledgements}

The authors would like to thank the members of NAVER Clova for proofreading this paper.
This work was supported by Institute of Information \& communications Technology Planning \& Evaluation (IITP) grant funded by the Korea government (MSIT)   (No. 2019-0-00075, Artificial Intelligence Graduate School Program (KAIST)).

\bibliography{anthology,custom}
\bibliographystyle{acl_natbib}

\newpage
\appendix
\section{Appendix}\label{sec:appendix}

\subsection{Training Details of the Passage Filter}\label{appendix:filter}
$J_\text{total}$ is the set of all 21M passages that serve as the knowledge source of DPR. $J^+_\text{train}$ consists of the top-200 passages retrieved for each of the questions in Natural Questions~\citep{nq} train, dev, test set, and EfficientQA dev set. To retrieved the passages, we use the retriever of~\citet{rdr} trained on Natural Questions. Smoothed frequency $\floor{\log_{10}(\text{freq}_{p_j}) + 1}$ is considered to create a candidate pool of the positive passages, and oversampling from the pool is done to make $J^{+'}_\text{train}$, whereas $J^{-'}_\text{train}$ is randomly and uniformly sampled from $J^{-}_\text{train}$. The objective function is defined as follows:
\begin{equation*}
\label{eqn:filter_obj}
    - \frac{\sum_{j \in J^{+'}_\text{train}} \log{\sigma(s_j)}
            +\sum_{j \in J^{-'}_\text{train}} \log{[1 - \sigma(s_j)}]}{ |J^{+'}_\text{train}| + |J^{-'}_\text{train}|}. \\
\end{equation*}
We finetune a RoBERTa-base~\citep{roberta} classifier with a batch size of 18 ($|J^{+'}_\text{train}| = |J^{-'}_\text{train}| = 9$), learning rate of 1e-5, dropout rate of 0.1, max norm gradient clipping of 2.0, and warmup steps of 1000, using one V100 GPU. We use the code of HuggingFace Transformers~\citep{transformers}, and no additional preprocessing is used on the data other than the tokenization for RoBERTa-base. We train the model for one epoch until all the positive passages oversampled according to the smoothed frequency are seen by the model at least once.

We evaluate the model on the validation set after every 2000 steps of gradient update.
We compose $J^+_\text{val}$ as a set of indices of the passages that hold the answer for one of the questions in the EfficientQA dev set and are retrieved by an existing retriever as the most relevant passage to the question. We use $|J^+_\text{val}| = 893$ positive passages and $|J^-_\text{val}| = 10000 - 893 = 9107$ randomly selected negative passages. $n = 1,224,000$ passages are selected for use in our minimal retrieve \& read system to fit in the storage budget of 500MB.

\subsection{Training Details of the Retriever-Reader with Single Encoder}\label{appendix:sharing}

\begin{table}[t]
    \setlength{\tabcolsep}{3.5pt}
    \centering
    \caption[Caption]{\small Statistics of the number of questions in each dataset. The values in parenthesis denote the number of questions filtered by DPR preprocessing and used for actual training. There is no training data for EfficientQA, so we use the training set of Natural Questions to train the model for EfficientQA.}
    \label{table:datasets}
    \footnotesize
    \begin{threeparttable}
    \begin{tabular*}{0.5\textwidth}{l@{\extracolsep{\fill}}rrr}
        \toprule
\textbf{Dataset}                     & \textbf{Train} & \textbf{Dev} & \textbf{Test} \\
\midrule
EfficientQA         & - & 1,800 & 1,800 \\
Natural Questions   & 79,168 (58,880) & 8,757 & 3,610        \\
TriviaQA            & 78,785 (60,413) & 8,837 & 11,313       \\
        \bottomrule
        \vspace*{-.5cm}
    \end{tabular*}
    \end{threeparttable}
\end{table}

We have not searched for hyperparameters in almost all experiments on parameter sharing and mainly followed the training setup of DPR.\footnote{\href{https://github.com/facebookresearch/DPR}{https://github.com/facebookresearch/DPR}}

For the experiments on EfficientQA, the training set of Natural Questions~\citep{nq} is used to train the models. The checkpoints that report the best result on the EfficientQA dev set\footnote{\href{https://github.com/google-research-datasets/natural-questions/blob/master/nq_open/NQ-open.efficientqa.dev.1.1.jsonl}{https://github.com/google-research-datasets/natural-\\questions/blob/master/nq\_open/NQ-open.efficientqa.dev.\\1.1.jsonl}} are selected. Our code is built on top of the official implementation of DPR, so the datasets are preprocessed as done in the code of DPR. Table~\ref{table:datasets} shows the statistics of the datasets including Natural Questions and TriviaQA used for the experiments in Appendix~\ref{appendix:nq_tqa}. We train the models using four to eight P40 or V100 GPUs.

\paragraph{Retriever Encoder Sharing}
We use MobileBERT~\citep{mobilebert} as the pretrained encoder. The encoder output vector is what corresponds to the \texttt{[CLS]} token in the input, and only the first 128 out of 512 dimensions are utilized to calculate the retrieval score. Following the setup of \citet{dpr}, we use a learning rate of 2e-5, max norm gradient clipping of 2.0, warmup of 1237, sequence length of 256, in-batch negative training of 1 positive and 127 negatives, and training epochs of 40 to 70, applying early stopping if there is a lack of resource. The models are evaluated on the dev set after every epoch.

\paragraph{Unified Retriever-Reader Through Knowledge Distillation}
$k' = 200$ and $m = 24$ is used to create the training dataset.
To train the unified retriever-reader, we use a learning rate of 1e-5, max norm gradient clipping of 2.0, no warmup steps, sequence length of 350, batch size of 16, knowledge distillation temperature $\tau$ of 3, and training epochs of 16 to 30, applying early stopping when the score seems to be converged.

Since the reader teacher (DPR reader) has a ranker, $\mathcal{L}_\text{read}$ is defined as the sum of the KL divergence between the span scores and the KL divergence between the ranking scores of the teacher and the student. $\mathcal{L}_\text{ret}$ also takes the passage ranking score from the ranker as the distillation target.

The model is evaluated at every 2000 steps, and we select the checkpoint with the highest average EM on $m'$ retrieved passages, where $m' \in \{1, 10, 20, 30, 40, 50\}$, along with an acceptable reranking accuracy (how many times the positive passage is ranked at the top among 50 candidates).

\paragraph{Iterative Finetuning of Unified Retriever-Reader}
We finetune the model for at most six epochs. The rest of the hyperparameters are the same as described in the previous paragraphs.

\subsection{Evaluation Details}\label{appendix:eval}

The reported EM is the highest EM on $m'$ retrieved passages where $m' \in \{1, 10, 20, \cdots, 100\}$.
The original code of DPR searches the answer only in the passage scored the highest by the passage ranker, and thus the answer span with the highest span score in the single passage is selected as the final answer. All of the EM scores presented in this work are also calculated this way.

On the other hand, we have found out that the end-to-end QA accuracy can be slightly increased by using the weighted sum of the passage ranking score $\mathcal{P}_\text{rank}$ and the answer span scores, $\mathcal{P}_\text{start}$ and $\mathcal{P}_\text{end}$ for the start and end positions, respectively, to compare the answer candidates at inference time. Therefore, we have used this scoring method for the model submitted to the EfficientQA leaderboard. Specifically, we use $(1 - \lambda) (\log{\mathcal{P}_\text{start}} + \log{\mathcal{P}_\text{end}}) + 2\lambda \log{\mathcal{P}_\text{rank}}$ as the score. The answer spans with the top five weighted sum scores in each retrieved passage are selected as the candidate answers, and the one with the highest score is chosen as the final answer. We select $\lambda=0.8$ based on the performance on the dev set. This method increases the dev set accuracy after the iterative finetuning stage (\S 2.2.3) from 34.44 to 34.61.

Due to the discrepancy between the validation accuracy during and after training (described in detail in Appendix~\ref{appendix:nq_tqa}), we select up to five checkpoints based on the dev set accuracy observed during training and evaluate them to obtain the one with the actual highest dev set accuracy after the iterative finetuning is done.

\subsection{System Footprint}\label{appendix:ablations}

\begin{table*}[ht]
\centering
    \setlength{\tabcolsep}{3.5pt}
    \captionsetup{width=.8\textwidth}
    \caption{\small Detailed ablations on how the system size changes from 77.5 GB to 484.69 MB by applying each of the discussed methods from the top to bottom. The values in the table use MB as the unit. Decreased values are marked in red and increased values are marked in blue.}
    \label{table:ablations}
    \footnotesize
    \begin{threeparttable}
    \begin{tabular*}{0.98\textwidth}{l@{\extracolsep{\fill}}rrrrrr}
        \toprule
& \textbf{Docker} & \textbf{Retriever} & \textbf{Reader} & \textbf{Index} & \textbf{Text File} & \textbf{Total} \\
\midrule
DPR                                                 & 1,270   & 836       & 418    & 61,919 & 13,065     & 77,508 \\
\S 2.1 Passage Filtering (21,015,325 $\rightarrow$ 1,224,000 passages)                               & 1,270   & 836       & 418    & {\color{red}3,681}  & {\color{red}756}       & {\color{red}6,961}  \\
\S 2.2.1 Embedding Dimension Reduction (768D $\rightarrow$ 128D)                                          & 1,270   & 836       & 418    & {\color{red}614}   & 756       & {\color{red}3,894}  \\
\S 2.2.1 Lightweight Encoder (BERT $\rightarrow$ MobileBERT)                                    & 1,270   & {\color{red}188}       & {\color{red}94}     & 614   & 756       & {\color{red}2,922}  \\
\S 2.2.2 Retriever Encoder Sharing               & 1,270   & {\color{red}94}        & 94     & 614   & 756       & {\color{red}2,828}  \\
\S 2.2.3 Unified Retriever-Reader Through Knowlege Distillation & 1,270   & 94        & {\color{red}0}      & 614   & 756       & {\color{red}2,734}  \\
\S 2.2.4 Iterative Finetuning of Unified Retriever-Reader                          & 1,270   & 94        & 0      & 614   & 756       & 2,734  \\
\S 2.3 INT8 Quantization of Index Items            & 1,270   & 94        & 0      & {\color{red}170}   & 756       & {\color{red}2,290}  \\
\S 2.3 Saving Model Weights as FP16                              & 1,270   & {\color{red}47}        & 0      & 170   & 756       & {\color{red}2,243}  \\
\S 2.3 Resource Compression                            & 1,270   & {\color{red}42}        & 0      & {\color{red}145}   & {\color{red}187}       & {\color{red}1,644}  \\
\S 2.3 Utilizing Token IDs as the Corpus                          & 1,270   & 42        & 0      & 145   & {\color{red}177}       & {\color{red}1,634}  \\
\makecell[l]{\S 3 TF Serving, Minimizing Library Dependencies, \\ \quad Fusing Index into Model Graph}                    & {\color{red}312}    & {\color{blue}177}       & 0      & {\color{red}0}     & 177       & {\color{red}666}   \\
\S 2.3 System Resource Compression                       & {\color{red}130}    & 177       & 0      & 0     & 177       & \textbf{\color{red}484}  \\
        \bottomrule
    \end{tabular*}
    \end{threeparttable}
\end{table*}

System footprint is measured by the command \texttt{du -h /} inside the standalone docker container right after its launching as stated in the EfficientQA competition guideline. The system footprint at runtime may be larger when the resources are initially compressed at the time of launching the container.

Table~\ref{table:ablations} shows from the top to bottom the detailed ablations on how the system size changes from 77.5 GB to 484.69 MB by applying each of the methods discussed in Section~\ref{sec:method}. The values in the table use MB as the unit. Decreased values are marked in red and increased values are marked in blue.

The docker image is initially assumed to be bitnami/pytorch:1.4.0\footnote{\href{https://hub.docker.com/r/bitnami/pytorch}{https://hub.docker.com/r/bitnami/pytorch}}, and it changes to python:3.6.11-slim-buster\footnote{\href{https://hub.docker.com/_/python}{https://hub.docker.com/\_/python}} after adopting TensorFlow (TF) Serving that does not require heavy system libraries as PyTorch does. The most lightweight docker image with python uses Alpine, but TF Serving does not run on an Alpine docker container due to the lack of support of system library requirements.

\subsection{Experiments: NQ and TriviaQA}\label{appendix:nq_tqa}

\begin{figure*}[ht!]
\centering
\begin{subfigure}[b]{0.48\textwidth}
\caption{\small Natural Questions}
\includegraphics[width=\textwidth]{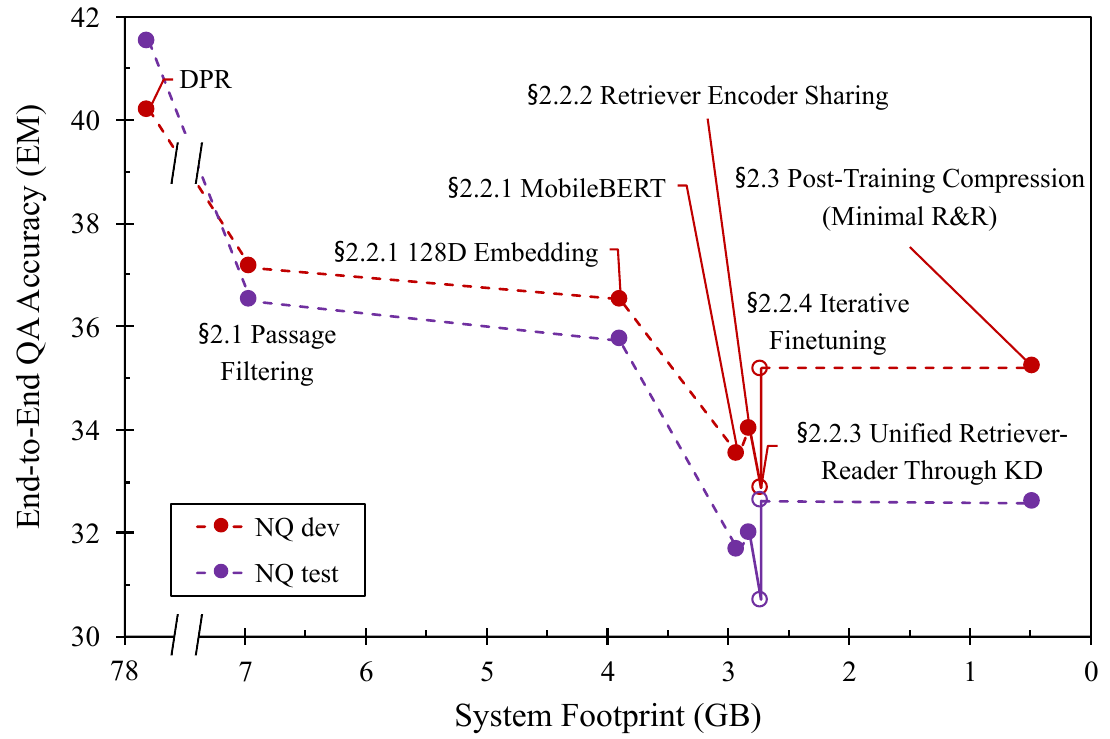}
\end{subfigure}
\begin{subfigure}[b]{0.48\textwidth}
\caption{\small TriviaQA}
\includegraphics[width=\textwidth]{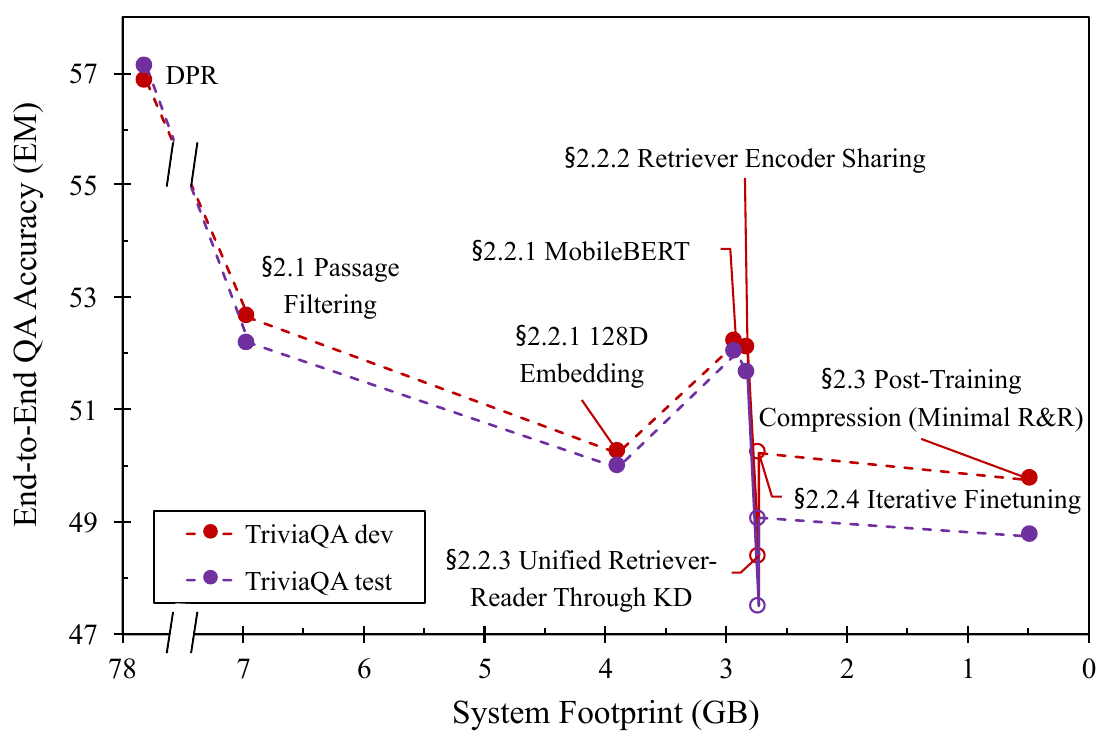}
\end{subfigure}
\caption{\small System footprint vs. Exact Match (EM) accuracy on Natural Questions and TriviaQA.}
\label{fig:nq_tqa}
\end{figure*}

\begin{table*}[ht]
\centering
    \setlength{\tabcolsep}{3.5pt}
    \captionsetup{width=.8\textwidth}
    \caption{\small Detailed ablations on how the Exact Match (EM) score on each dataset changes by applying each of the discussed methods from the top to bottom.}
    \label{table:overall}
    \footnotesize
    \begin{threeparttable}
    \begin{tabular*}{0.98\textwidth}{l@{\extracolsep{\fill}}rrrrr}
        \toprule
                     & \textbf{EfficientQA Dev} & \textbf{NQ Dev} & \textbf{NQ Test} & \textbf{TriviaQA Dev} & \textbf{TriviaQA Test} \\
        \midrule
DPR                                         & 36.78           & 40.20  & 41.52   & 56.84        & 57.10         \\
§2.1 Passage Filtering                      & 34.06           & 37.14  & 36.51   & 52.65        & 52.19         \\
§2.2.1 128D Embedding                       & 32.67           & 36.52  & 35.73   & 50.23        & 49.97         \\
§2.2.1 MobileBERT                           & 32.61           & 33.52  & 31.66   & 52.20        & 52.02         \\
§2.2.2 Retriever Encoder Sharing            & 32.78           & 34.03  & 31.99   & 52.11        & 51.66         \\
§2.2.3 Unified Retriever-Reader Through KD  & 31.67           & 32.88  & 30.72   & 48.40        & 47.51         \\
§2.2.4 Iterative Finetuning                 & 34.44           & 35.19  & 32.63   & 50.23        & 49.06         \\
§2.3 Post-Training Compression              & 34.33           & 35.22    & 32.60     & 49.76          & 48.75    \\
        \bottomrule
    \end{tabular*}
    \end{threeparttable}
\end{table*}

\paragraph{Experimental Setup}
Most of the details to train the models on Natural Questions (NQ) and TriviaQA (Trivia) follow what is written in Appendix~\ref{appendix:filter} and Appendix~\ref{appendix:sharing}, and here we describe only the differences.
To train the passage filter, we use log with base 2 instead of 10 for Trivia due to its higher validation set accuracy. The questions used to create the training data are from the train and dev set of the datasets which correspond to the targets of the filter models.
To train the unified retriever-reader through knowledge distillation, a batch size of 8 with gradient accumulation steps of 2 is used to train the models using only four V100 GPUs. The maximum number of training epochs is set to 30, but training is stopped around the 16th epoch to shorten the training time even when the scores do not seem to be fully converged.
For iterative finetuning, a batch size of 8 with gradient accumulation steps of 2 is again used with four V100 GPUs. The maximum number of training epochs is also set to 30, but the training is stopped before the 10th epoch.

\paragraph{Experimental Results}
Figure~\ref{fig:nq_tqa} shows the EM and docker-level system footprint when each of the discussed strategies is applied to DPR.
In the case of the EfficientQA dataset, the step-wise evaluation result on the test set cannot be reported because the answer set is not publicly available. On the other hand, for NQ and Trivia, we present the step-wise accuracy on the test set along with that on the dev set to show how the strategies affect the actual performance on the test set. The evaluation results of all the cases are presented in Table~\ref{table:overall}.

Let us define the relative performance drop at step $t$ as the percentage of $\frac{\text{EM}_{t-1} - \text{EM}_{t}}{\text{EM}_{t-1}}$ where $\text{EM}_{t}$ is the EM score at the $t$-th phase.
As shown in the figures and the table, applying the methods to different datasets does not show consistent trends. Because the EfficientQA dataset is constructed in the same way as NQ~\citep{efficientqa}, the trends on these two datasets are similar except that changing the backbone from BERT to MobileBERT (\S 2.2.1) results in a significant relative performance drop of 8.21\% on the dev set of NQ while the value is only 0.18\% on EfficientQA. On the other hand, the same change results in about 4\% of relative performance \emph{gain} on Trivia. A different phenomenon also appears when the retriever encoders are shared (\S 2.2.2) that the accuracy rises on EfficientQA and NQ while it drops on Trivia.

The percentage of the final accuracy to the accuracy at the start also differs among the datasets: 93.3\% and 89.0\%\footnote{36.0 is used as an approximation for the accuracy of DPR on the EfficientQA test set, which is reported as 36 in \href{https://github.com/google-research-datasets/natural-questions/tree/master/nq_open}{https://github.com/google-research-datasets/natural-questions/tree/master/nq\_open}.} on the EfficientQA dev and test set, 87.6\%, 87.5\%, and 85.4\% on the NQ dev set, Trivia dev set, and Trivia test set, respectively, but 78.5\% on the NQ test set. While the gap between the percentages on the dev and test set is small on Trivia, the value is considerably large on NQ. Also, the gap between the dev and test set accuracy divided by the latter is about 7\% on EfficientQA and NQ, while it is only 2\% on Trivia.

Meanwhile, a common observation is that passage filtering (\S 2.1), embedding dimension reduction (\S 2.2.1), and unifying the retriever and the reader through knowledge distillation (\S 2.2.3) consistently result in the drop of accuracy. The relative performance drop of each of the methods is 7.40\%, 4.08\%, and 3.39\% on the EfficientQA dev set, 7.61\%, 1.67\%, and 3.38\% on the NQ dev set, 12.07\%, 2.14\%, and 3.97\% on the NQ test set, 7.37\%, 4.60\%, 7.12\% on the Trivia dev set, and 8.60\%, 4.25\%, 8.03\% on the Trivia test set.\footnote{Figure 1 of \citet{memory} also shows the trade-off between the index size and system accuracy. Note that the implementation details of their passage filtering and embedding dimension reduction are different from ours.}

In the case of unifying the retriever and the reader into one model, one possible cause of the accuracy drop might have come from its currently suboptimal checkpoint selection method. From the moment the retriever and reader are unified into one model and jointly trained, the validation accuracy reported during training uses the outputs of the \textit{initial} retriever parameters while the actual evaluation must use outputs of the \textit{updated} retriever parameters at the time of validation. Due to this discrepancy, checkpoint selection based on the validation accuracy at training does not lead to the model with the actual highest dev set accuracy. The discrepancy may further necessitate measuring the true dev set accuracy at several different checkpoints (possibly with high validation accuracy during training) to choose the final model after iterative finetuning. To deal with this issue and fairly compare the best checkpoints, future research may be conducted to refresh the retrieval index during training as in the work of~\citet{realm,ance}, so that the evaluation (and training) may not be done on the stale retrieval outputs.

\end{document}